\documentclass[11pt,a4paper]{article}

\usepackage[margin=1in]{geometry}
\usepackage[utf8]{inputenc}
\usepackage[numbers,sort&compress]{natbib}
\usepackage{authblk}
\usepackage{graphicx} 
\usepackage{url}
\usepackage{booktabs}
\usepackage{float}
\usepackage{tabularx}
\usepackage{pdflscape} 
\usepackage{multirow}
\usepackage{longtable}
\usepackage{glossaries}

\newacronym{adamw}{AdamW}{Adam Optimizer with Weight Decay}
\newacronym{auc}{AUC}{Area Under the Curve}
\newacronym{cie}{CIE}{Clasificación Internacional de Enfermedades}
\newacronym{eda}{EDA}{Exploratory Data Analysis}
\newacronym{f1}{F1}{F1-score}
\newacronym{gpais}{GPAIS}{Sistema de Inteligencia Artificial de Propósito General}
\newacronym{icd}{ICD}{International Classification of Diseases}
\newacronym{llm}{LLM}{Large-Language Model}
\newacronym{lora}{LoRA}{Low-Rank Adaptation}
\newacronym{ml}{ML}{Machine Learning}
\newacronym{nf4}{NF4}{Normalized Float-4 Quantization}
\newacronym{nlp}{NLP}{Natural Language Processing}
\newacronym{ovr}{OvR}{One-vs-Rest}
\newacronym{qlora}{QLoRA}{Quantized Low-Rank Adaptation}
\newacronym{rag}{RAG}{Retrieval-Augmented Generation}
\newacronym{relu}{ReLU}{Rectified Linear Unit}
\newacronym{roc}{ROC}{Receiver Operating Characteristic}
\newacronym{upm}{UPM}{Universidad Politécnica de Madrid}
\newacronym{vllm}{vLLM}{Versatile Large Language-Model serving framework}
\newacronym{vram}{VRAM}{Video Random Access Memory}
\newacronym{xgb}{XGBoost}{Extreme Gradient Boosting}
\newacronym{cnn}{CNN}{Convolutional Neural Network}
\newacronym{lsa}{LSA}{Latent Semantic Analysis}
\newacronym{lda}{LDA}{Latent Dirichlet Allocation}
\newacronym{tpe}{TPE}{Tree-structured Parzen Estimator}
\newacronym{bow}{BoW}{Bag of Words}
\newacronym{tfidf}{TF-IDF}{Term Frequency-Inverse Document Frequency}
\newacronym{dl}{DL}{Deep Learning}
\newacronym{mlp}{MLP}{Multi Layer Perceptron}

\usepackage{cleveref}

\title{Automated ICD Classification of Psychiatric Diagnoses: From Classical NLP to Large Language Models}

\author[1,2,*]{Fernando Ortega}
\author[1,2]{Raúl Lara-Cabrera}
\author[1,2]{Jorge Dueñas-Lerín}
\author[10,11]{Alejandro {de la Torre-Luque}}
\author[3]{Mercé Salvador Robert}
\author[4,5,6,7,8,9,10]{Enrique Baca-García}

\affil[1]{\small Department of Sistemas Informáticos, Universidad Politécnica de Madrid, Spain}
\affil[2]{\small KNODIS Research Group, Universidad Politécnica de Madrid, Spain}
\affil[3]{\small Hospital Universitario de Móstoles, Universidad Rey Juan Carlos, Spain}
\affil[4]{\small Department of Psychiatry, University Hospital Jimenez Díaz Fundation, Madrid, Spain}
\affil[5]{\small Department of Psychiatry, University Hospital Rey Juan Carlos, Móstoles, Spain}
\affil[6]{\small Department of Psychiatry, General Hospital of Villalba, Madrid, Spain}
\affil[7]{\small Department of Psychiatry, University Hospital Infanta Elena, Madrid, Spain}
\affil[8]{\small Department of Psychology, Universidad Catolica del Maule, Talca, Chile}
\affil[9]{\small Department of Psychiatry, Madrid Autonomous University, Madrid, Spain}
\affil[10]{\small CIBERSAM ISCIII, Spain}
\affil[11]{\small Department of Legal Medicine, Psychiatry and Pathology. Complutense University of Madrid, Spain}
\affil[*]{\small Corresponding author: fernando.ortega@upm.es}

\date{}

\begin{document}

\maketitle

\begin{abstract}
Mental health has become a global priority, leading to a massive administrative burden in the coding of clinical diagnoses. This study proposes the automation of psychiatric diagnostic analysis by mapping free-text descriptions to the International Classification of Diseases (ICD) using Natural Language Processing (NLP) and Machine Learning (ML) techniques. Utilizing a specialized dataset of 145,513 Spanish psychiatric descriptions, various text representation paradigms were evaluated, ranging from classical frequency-based models (BoW, TF-IDF) to state-of-the-art Large Language Models (LLMs) such as e5\_large, BioLORD, and Llama-3-8B. Results indicate that transformer-based embeddings consistently outperform traditional methods by capturing implicit semantic cues and nuanced medical terminology. The e5\_large model, through end-to-end fine-tuning, achieved the highest performance with a $F1_{micro}$ score of 0.866. This research demonstrates that adapting LLMs to specific clinical nomenclature is essential for overcoming the challenges of ``long-tail'' label distributions and the inherent ambiguity of psychiatric discourse.
\end{abstract}

\vspace{1em}
\vspace{1em}

\section*{Research Highlights}
\begin{itemize}
    \item \textbf{Large-scale Spanish Clinical Dataset:} The study leverages a massive real-world dataset of over 145,000 mental health diagnostic descriptions.
    \item \textbf{Semantic Depth vs. Keyword Matching:} Transformer-based models (specifically e5\_large and BioLORD) significantly outperformed classical BoW/TF-IDF approaches by understanding clinical context rather than relying on explicit keywords.
    \item \textbf{Optimal Classifier Synergies:} Research found that XGBoost serves as the most robust head for dense contextual embeddings, while Multi-Layer Perceptrons (MLP) are uniquely effective for high-dimensional sparse data.
    \item \textbf{Superiority of Fine-Tuning:} Transitioning from static feature extraction to active end-to-end fine-tuning of LLMs yielded the study's peak performance, reaching a 0.866 F1 micro score.
    \item \textbf{Addressing the Long-Tail Challenge:} The study highlights the persistent difficulty of coding low-prevalence psychiatric conditions, where semantic richness alone cannot fully compensate for extreme class imbalance.
\end{itemize}

\section{Introduction} \label{sec:introduction}

Mental health has become a public health priority: nearly 34\% of the Spanish population reported having suffered some psychological problem during 2023, with peaks exceeding 40\% in those over 50 years old~\cite{informe_sns_2023}. Anxiety disorders, for example, affect 106.5 people per thousand inhabitants~\cite{salud_mental_aldama_2024}, and their socioeconomic impact is reflected in a sustained increase in the consumption of psychotropic drugs and healthcare costs~\cite{pastillas_pais_2025}. Previous psychiatric diagnoses are strong predictors of the subsequent development of additional mental disorders. Several longitudinal and epidemiological studies have reported that between 15\% and 50\% of individuals diagnosed with a mental disorder may later develop another psychiatric condition, highlighting the high prevalence of psychiatric comorbidity and diagnostic progression \cite{McGrath2020,Hamad2025Dec}.

In this context, the automatic exploration and classification of diagnostic information from clinical records could support the early identification of patients at risk of future psychiatric episodes. In this paper, we propose automating diagnostic analysis tasks using \gls{nlp} and \gls{ml} techniques to alleviate administrative burden, improve coding consistency and accuracy, and reduce both processing time and errors. Beyond facilitating clinical coding workflows, automated \gls{icd} classification systems may also optimize healthcare resources and enable predictive analytics for identifying future psychiatric trajectories and comorbid conditions.

The remainder of this paper is structured as follows. \Cref{sec:state-of-the-art} reviews current state-of-the-art methods for automating the diagnosis of mental health disorders. \Cref{sec:material-and-methods} details the materials and methods, describing the Spanish clinical dataset, the preprocessing pipeline, and the text representation techniques evaluated. \Cref{sec:results} presents the experimental setup, the evaluation metrics, and a comparative analysis of the classification performance for \gls{icd} codes. \Cref{sec:discussion} provides an in-depth discussion of the findings, focusing on the challenges of Spanish medical language and the trade-off between model interpretability and predictive accuracy. Finally, \cref{sec:conclusions} summarizes the main contributions and outlines potential directions for future research.

\section{State of the art} \label{sec:state-of-the-art}

Automated clinical coding involves mapping natural language diagnostic descriptions to standardized codes from healthcare classifications, such as \gls{icd}. This process is fundamental for data analysis, clinical research, and the systematic collection of health statistics. Traditionally, this task has been performed manually by subject-matter experts, making it laborious, costly, and susceptible to human error~\cite{hou2025enhancing}. This challenge is particularly acute in fields such as psychiatry and mental health, where clinical notes often contain subjective descriptions of symptoms and contexts that may not explicitly reference the formal name of a disorder~\cite{silva2024aiding}.

In recent years, the field has advanced significantly due to the emergence of \gls{nlp} and \glspl{llm}. Automatic clinical code assignment is framed as an extreme multi-label classification problem, where a single note may be associated with dozens of codes from a label space of thousands of \gls{icd} entries. Historically, an \gls{ovr} scheme was employed, training a binary classifier for each individual code to predict its presence. However, a significant breakthrough occurred with the introduction of CAML~\cite{mullenbach2018explainable}. This model utilizes a \gls{cnn} to extract text representations alongside a label-specific attention mechanism, allowing each code to identify the most relevant phrases within the clinical text. 

Subsequent refinements have leveraged the inherent hierarchy and relationships between codes. For instance, LAAT~\cite{vu2020label} introduced multiple layers of label-wise attention and modeled dependencies between labels. Concurrently, models based on graphs and label \emph{embeddings} emerged: MSATT-KG~\cite{xie2019msattkg} combines a \gls{cnn} with a graph network to propagate information through the medical ontology hierarchy, while HyperCore~\cite{hypercore} projects codes into a hyperbolic space and employs convolutions over co-occurrence graphs to improve the representation of infrequent labels.

More recently, research has explored reformulating the task as a sequential label generation problem. For example, \cite{yang2022multilabel} treated the process as an autoregressive generation task to capture implicit dependencies, showing competitiveness in low-data scenarios, although it requires mechanisms to ensure that the generated codes are valid. Furthermore, the Transformer architecture has inspired extreme classifiers such as XR-Transformer~\cite{zhang2021fast} and its hierarchical variant XR-LAT~\cite{liu2022xrlat}, which perform predictions across different levels of the code tree.

The rise of \glspl{llm} (such as GPT-3, GPT-4, PaLM, LLaMA, or BERT) offers a promising new paradigm for automatic coding. However, using these models in their ``off-the-shelf'' form often yields suboptimal results~\cite{hou2025enhancing} due to code hallucinations, confusion between clinically similar diagnoses, and poor coverage of rare codes. A common strategy involves adapting pre-trained \glspl{llm} to the medical domain by training them on large biomedical or clinical datasets. Notable pioneers include \emph{BioBERT}~\cite{lee-etal-2020-biobert}, trained on biomedical publications, and \emph{ClinicalBERT}~\cite{alsentzer-etal-2019-clinicalbert}, which incorporates clinical notes. These variants demonstrate a deeper understanding of medical nuance compared to generic \emph{BERT} models. Other models such as \emph{BlueBERT}~\cite{peng2019bluebert} and \emph{PubMedBERT}~\cite{gu2021pubmedbert} have further refined these domain-specific representations.

In the multilingual context, specifically for Spanish, Carrino~\cite{carrino-etal-2022-pretrained} introduced the first biomedical \emph{BERT} models trained from scratch on 1.1 billion words, proving that language-specific domain adaptation outperforms generic models. Similarly, \emph{MédicoBERT}~\cite{padilla2024medicobert} was developed to adapt \emph{BERT} to the Spanish medical lexicon for comprehension and \textit{question-answering} tasks.

These pre-trained models provide high-quality \emph{embeddings} that serve as the foundation for various downstream tasks. The standard practice involves fine-tuning these models on annotated clinical datasets, as seen in PLM-ICD~\cite{huang-etal-2022-plm}. This approach has been shown to substantially outperform traditional bag-of-words or networks trained from scratch, as it better captures the subtleties of medical discourse. \citet{hou2025enhancing} proposed a two-phase approach: first, training the model on the definitions of 74,260 \gls{icd} codes, followed by fine-tuning with real clinical notes containing abbreviations and typographical errors. Their LLaMA-based version significantly outperformed un-tuned GPT-3.5 models, demonstrating that specialized \glspl{llm} can approach human-expert performance even in complex scenarios. This success is attributed to the integration of structured knowledge (official definitions) with exposure to the diversity of real-world clinical language. 

Finally, another research direction combines these approaches by utilizing the \gls{llm} as a component within a broader system. LLM-Codex~\cite{yang2023llmcodex} presents a two-step pipeline where the \gls{llm} generates \texttt{<code, textual evidence>} pairs with high recall, while an external verifier filters false positives by cross-referencing each prediction against the source note.

\section{Material and Methods} \label{sec:material-and-methods}

\begin{figure}[ht]
\centering
\includegraphics[width=0.9\textwidth]{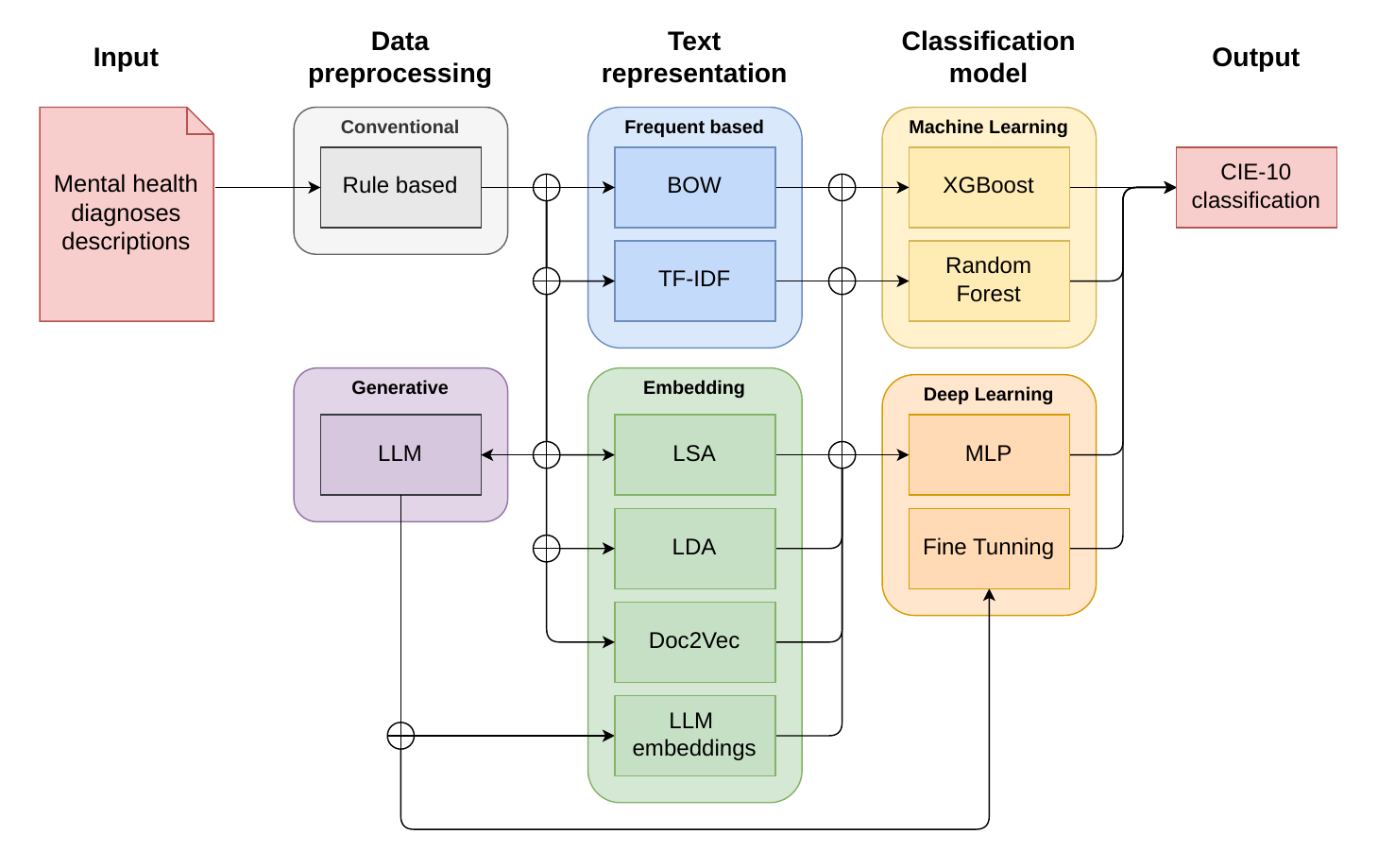}
\caption{Workflow of the proposed methodology, from raw input to \gls{icd} classification.}
\label{fig:summary}
\end{figure}

This section details the methodology employed to evaluate the impact of different text representation techniques on the automated assignment of \gls{icd} codes. \Cref{fig:summary} summarizes the pipeline used to evaluate various approaches for classifying mental health diagnostic descriptions into \gls{icd} codes.

The \textbf{input data} for this research comprises 145,513 descriptions in Spanish authored by healthcare professionals regarding mental health diagnoses. These entries consist of free-text fields; consequently, the style, length, and phrasing vary significantly depending on the individual professional. Based on these descriptions, a panel of experts classified each entry into one or more codes from a selection of 85 standardized diagnoses. This resulted in a subset of 79,048 entries that include both the original medical description and the corresponding expected standardized outcome (ground truth). Furthermore, the set of 85 \gls{icd} codes selected for this specific purpose can be seen in \cref{apx:cie-10-codes}, most of which include their respective textual labels. These codes constitute the available categories for the diagnostic standardization process. 

Regarding ethical considerations, the data were received in an anonymized format. The information provided was limited strictly to the diagnostic texts and their associated standards, excluding any personal identifiers or references that could be linked to the patient. The data for this research were provided by the \textit{Hospital Fundación Jiménez Díaz, Grupo Hospitalario Quirón}.

\begin{figure}[ht]
\centering
\includegraphics[width=0.9\textwidth]{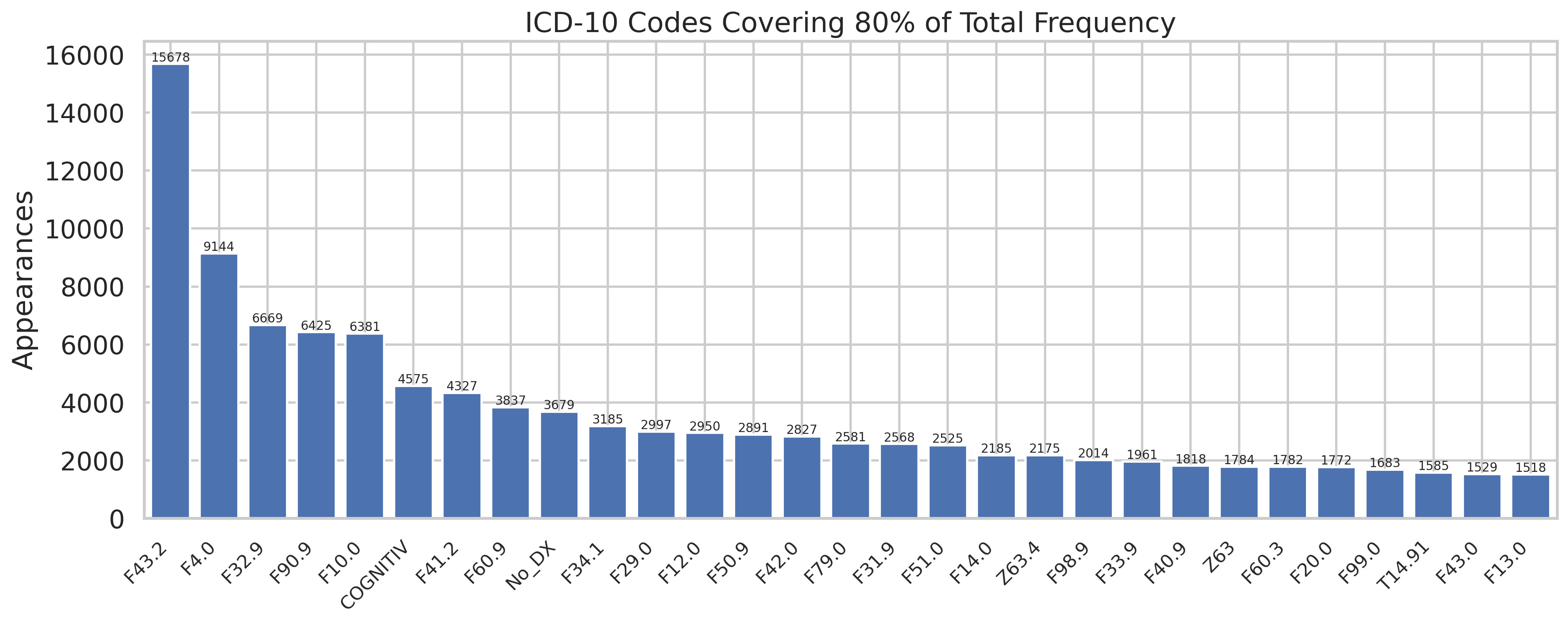}
\caption{\gls{icd} code frequencies. As illustrated, 29 \gls{icd} codes represent 80\% of total appearances.}
\label{fig:icd10_frequencies}
\end{figure}

\Cref{fig:icd10_frequencies} shows the \gls{icd} code frequencies. The labels exhibit a long-tail distribution and the severe class imbalance characteristic of medical coding datasets.

The \textbf{data preprocessing} phase was designed to mitigate the inherent noise of free-text psychiatric clinical notes. The core pipeline began with a text enrichment step: regular expressions were used to identify abbreviated \gls{icd} codes, which were then replaced with their full, standardized textual descriptions to expand the records' semantic context. Subsequently, rule-based methods were applied to standardize formats and remove non-informative characters.

Following the preprocessing stage, the clinical notes must be transformed into numerical features suitable for a multi-label classification framework. We explore several \textbf{text representation} paradigms to identify the optimal semantic richness for psychiatric discourse. These techniques are categorized as follows:

\begin{itemize}
\item \textbf{Frequent-based Representations:} We implement the classical \gls{bow} model and its weighted variant, \acrfull{tfidf}. These methods represent each clinical note as a high-dimensional sparse vector, capturing the statistical importance of specific medical terms while dampening the influence of common linguistic noise.
\item \textbf{Embedding approaches:} To address the challenges of synonymy and medical jargon, we apply \gls{lsa}~\cite{dumais2004latent} and \gls{lda}~\cite{blei2003latent}. These dimensionality reduction and probabilistic modeling techniques allow us to extract latent thematic structures (topics) from the unstructured text, providing a more condensed semantic representation. Additionally, we utilize \texttt{Doc2Vec}~\cite{le2014distributed} to generate dense, fixed-length vectors. By training on the specific corpus of Spanish psychiatric notes, this model learns to represent entire documents in a continuous vector space, capturing contextual dependencies typically lost in bag-of-words approaches. Finally, we leverage state-of-the-art \glspl{llm} to extract deep contextualized embeddings. In this approach, the clinical text is passed directly through the transformer architecture of the models shown in \cref{tab:llms} to obtain the final hidden-state representations. These embeddings encode complex semantic relationships and nuanced medical terminology, serving as high-fidelity features for the subsequent classification task.
\end{itemize}

\begin{table}[ht]
\label{tab:llms}
\footnotesize
\caption{\glspl{llm} used to generate contextual embeddings for the clinical notes.}
\begin{tabular}{llc}
\toprule 
\textbf{Model Name} & \textbf{ShortName} & \textbf{Number of parameters} \\ \midrule
PlanTL-GOB-ES/roberta-base-biomedical-clinical-es \cite{carrino2021biomedical} & \texttt{plantl\_roberta} & 125M \\ \midrule
mrm8488/bert-base-spanish-wwm-cased-finetuned-spa-squad2-es & \texttt{bert\_spanish} & 110M \\ \midrule
PlanTL-GOB-ES/bsc-bio-es \cite{carrino-etal-2022-pretrained} & \texttt{bsc\_bio} & 125M \\ \midrule
intfloat/multilingual-e5-base \cite{wang2024multilingual} & \texttt{e5\_base} & 270M \\ \midrule
intfloat/multilingual-e5-large \cite{wang2024multilingual} & \texttt{e5\_large} & 550M \\ \midrule
sentence-transformers/paraphrase-multilingual-mpnet-base-v2 \cite{reimers-2019-sentence-bert} & \texttt{paraphrase\_multilangual} & 270M \\ \midrule
FremyCompany/BioLORD-2023-M \cite{remy-etal-2023-biolord} & \texttt{bio\_lord} & 270M \\ \midrule
Henrychur/MMed-Llama-3-8B \cite{qiu2024building} & \texttt{llama3\_8b} & 8B \\ \bottomrule
\end{tabular}
\end{table}

These distinct representation strategies allow for rigorous benchmarking of how different levels of semantic abstraction affect the precision of automated \gls{icd} code assignment.

Finally, the multi-label \textbf{classification task} was addressed through three distinct modeling strategies, ranging from traditional machine learning to deep neural architectures and transformer-based fine-tuning. Regarding Machine Learning methods, we implemented \textit{Random Forest} and \textit{XGBoost}. These models were trained using the various feature sets derived from the techniques previously described, ranging from sparse frequency-based vectors to dense contextual embeddings. We evaluated these classifiers in their native multi-output configurations, allowing the tree-based algorithms to process the 87 labels simultaneously and potentially capture inherent dependencies between different psychiatric diagnoses.

To explore the performance of \gls{dl}, an \gls{mlp} was specifically designed to handle the diverse nature of the input features. Unlike the ensemble models, the internal architecture of this network—including the number of hidden layers, neurons per layer, and dropout rates—was not fixed; instead, it was dynamically determined during hyperparameter optimization (see \cref{subsec:hyperparameters}) to identify the most efficient topology for each representation paradigm. The only constant element was the output stage: a final layer of 85 neurons with sigmoid activation functions, optimized via binary cross-entropy loss. This setup allows the network to learn complex non-linear mappings and adapt its depth to the specificities of the input vector, whether processing high-dimensional sparse data or low-dimensional dense embeddings.

Finally, we leveraged the power of \glspl{llm} through end-to-end Fine Tunning, representing a shift from using these models as static feature extractors to active classifiers. In this approach, we appended a task-specific classification head to the transformer backbone and updated the model's internal weights using our specialized Spanish psychiatric dataset. This process enables the model to adapt its deep linguistic representations specifically to the clinical nomenclature and the long-tail distribution of the 87 codes. This strategy represents the most computationally intensive yet semantically sophisticated tier of our classification framework, as it optimizes the entire representation and decision-making pipeline simultaneously.

\section{Experimental Setup and Results} \label{sec:results}

In this section, we describe the experimental framework and present a comparative analysis of the results obtained. First, we define the evaluation metrics tailored for extreme multi-label classification, which are essential for measuring performance in large \gls{icd} label spaces. Following this, we report the classification F1, precision and recall for each of the representation techniques. These results provide an empirical basis to determine which method best captures the semantic complexity of clinical language while maintaining robustness across frequent and rare diagnostic codes.

\subsection{Evaluation Metrics} 

The evaluation of automatic coding systems requires metrics capable of reflecting both overall accuracy and performance on rare labels. Let $L$ be the set of labels, where in our case $|L| = 85$. For each label $l \in L$, we denote $TP_l, FP_l,$ and $FN_l$ as the number of true positives, false positives, and false negatives, respectively. The precision ($P_l$) and recall ($R_l$) for an individual label are defined as:

\begin{equation}
    P_l = \frac{TP_l}{TP_l + FP_l}, \quad R_l = \frac{TP_l}{TP_l + FN_l}
\end{equation}

The standard is to report \textit{micro-F1}, which aggregates the contributions of all classes to compute the average metric. It calculates precision and recall by considering each document-code pair as an independent instance and, therefore, weights frequent codes more heavily:

\begin{equation}
    P_{micro} = \frac{\sum_{l \in L} TP_l}{\sum_{l \in L} (TP_l + FP_l)}, \quad R_{micro} = \frac{\sum_{l \in L} TP_l}{\sum_{l \in L} (TP_l + FN_l)}
\end{equation}

The \textit{micro-F1} is then the harmonic mean of these global values: 

\begin{equation} \label{eq:f1-micro}
F1_{micro} = 2 \cdot \frac{P_{micro} \cdot R_{micro}}{P_{micro} + R_{micro}}
\end{equation}

On the other hand, \textit{macro-F1} averages the F1-score per label, giving the same weight to highly and poorly represented codes:

\begin{equation}
    F1_{macro} = \frac{1}{|L|} \sum_{l \in L} F1_l, \quad \text{where } F1_l = 2 \cdot \frac{P_l \cdot R_l}{P_l + R_l}
\end{equation}

An improvement in \textit{macro-F1} usually indicates better handling of minority classes, as it treats the performance on a rare label with the same importance as a frequent one.

\subsection{Hyperparameter Optimization} \label{subsec:hyperparameters}

To systematically identify the most effective configurations for our diverse set of classification models, we employed Optuna~\cite{akiba2019optuna}, an advanced define-by-run hyperparameter optimization framework. Optuna automates the search process by dynamically constructing the search space and efficiently sampling hyperparameters using Bayesian optimization strategies, typically the \gls{tpe}. The optimization objective across all configurations was to maximize the $F1_{micro}$ score (see \cref{eq:f1-micro}) evaluated on a strictly isolated validation set. 

Given the long-tail distribution of the labels (see \cref{fig:icd10_frequencies}), a multilabel stratified shuffle split algorithm was employed to divide the dataset into training (70\%), validation (15\%), and test (15\%) sets. This iterative stratification guaranties that the marginal probability distribution of each label, including highly infrequent psychiatric disorders, is strictly preserved across all partitions, ensuring a robust and unbiased model evaluation.

As explained in \cref{sec:material-and-methods}, we evaluated several text representation techniques and classification models, each requiring specific hyperparameter tuning to maximize performance. \Cref{tab:text_representation_search_space} contains the hyperparameters tested for each representation method, such as those used in frequent-based approaches or standard embedding techniques. Notably, \gls{llm} embeddings do not have parameters to optimize during this stage, as they are utilized as static feature extractors that generate fixed contextual vectors based on pre-trained weights without architectural modifications.

On the other hand, \cref{tab:classification_models_search_space} details the search space for the machine learning and deep learning classifiers. This includes XGBoost, Random Forest, and the \gls{mlp}, for which the layers' configuration—including depth and width—is also dynamically optimized to identify the most efficient topology for each specific representation paradigm.

Finally, the fine-tuning of the \glspl{llm} involves a simultaneous optimization of both the representation and the classification head. To mitigate catastrophic forgetting, we adopt a \textit{linear probing then fine-tuning} (LP-FT) strategy~\cite{kumar2022fine}, applying separate learning rates for the classification head and the transformer backbone. Additionally, to address class imbalance in the multilabel setting, positive weights derived from the negative-to-positive label ratio are incorporated into the loss function, with a smoothing exponent (pos\_weight\_alpha) controlling the correction magnitude.

\begin{table}[ht]
\centering
\footnotesize
\caption{Hyperparameter search space for text representation techniques.}
\label{tab:text_representation_search_space}
\begin{tabular}{@{}llp{7.2cm}@{}}
\toprule
\textbf{Representation Method} & \textbf{Hyperparameter} & \textbf{Search Space} \\ \midrule

\multirow{3}{*}{\textbf{BoW}}
 & \texttt{max\_features} & Integer $\in [1000, 10000]$, step $= 500$ \\
 & \texttt{min\_df} & Integer $\in [1, 5]$ \\
 & \texttt{max\_df} & Float $\in [0.80, 0.99]$, step $= 0.01$ \\ \midrule

\multirow{3}{*}{\textbf{TF-IDF}}
 & \texttt{max\_features} & Integer $\in [1000, 10000]$, step $= 500$ \\
 & \texttt{min\_df} & Integer $\in [1, 5]$ \\
 & \texttt{max\_df} & Float $\in [0.80, 0.99]$, step $= 0.01$ \\ \midrule

\multirow{4}{*}{\textbf{LSA}}
 & \texttt{n\_components} & Integer $\in [50, 500]$, step $= 50$ \\
 & \texttt{max\_features} & Integer $\in [1000, 10000]$, step $= 500$ \\
 & \texttt{min\_df} & Integer $\in [1, 5]$ \\
 & \texttt{max\_df} & Float $\in [0.80, 0.99]$, step $= 0.01$ \\ \midrule

\multirow{5}{*}{\textbf{LDA}}
 & \texttt{n\_topics} & Integer $\in [10, 100]$, step $= 10$ \\
 & \texttt{max\_iter} & Integer $\in [10, 50]$, step $= 10$ \\
 & \texttt{max\_features} & Integer $\in [1000, 10000]$, step $= 500$ \\
 & \texttt{min\_df} & Integer $\in [1, 5]$ \\
 & \texttt{max\_df} & Float $\in [0.80, 0.99]$, step $= 0.01$ \\ \midrule

\multirow{3}{*}{\textbf{Doc2Vec}}
 & \texttt{vector\_size} & Integer $\in [100, 500]$, step $= 50$ \\
 & \texttt{min\_count} & Integer $\in [1, 5]$ \\
 & \texttt{epochs} & Integer $\in [20, 100]$, step $= 10$ \\ 
\bottomrule
\end{tabular}
\end{table}

\begin{table}[ht]
\centering
\footnotesize
\caption{Hyperparameter search space for Machine Learning and Deep Learning classification models.}
\label{tab:classification_models_search_space}
\begin{tabular}{@{}llp{6cm}@{}}
\toprule
\textbf{Classification Model} & \textbf{Hyperparameter} & \textbf{Search Space} \\ \midrule
\multirow{7}{*}{\textbf{Random Forest}} 
 & \texttt{n\_estimators} & Integer $\in [5, 50]$ \\
 & \texttt{max\_depth} & Integer $\in [5, 25]$ \\
 & \texttt{min\_samples\_split} & Integer $\in [2, 6]$ \\
 & \texttt{min\_samples\_leaf} & Integer $\in [1, 4]$ \\
 & \texttt{max\_features} & Categorical $\in \{\text{sqrt}, \text{log2}, \text{None}\}$ \\
 & \texttt{bootstrap} & Categorical $\in \{\text{True}, \text{False}\}$ \\
 & \texttt{class\_weight} & Categorical $\in \{\text{None}, \text{balanced}\}$ \\ \midrule
\multirow{9}{*}{\textbf{XGBoost}} 
 & \texttt{n\_estimators} & Integer $\in [100, 800]$ \\
 & \texttt{learning\_rate} & Log-Uniform Float $\in [10^{-3}, 0.3]$ \\
 & \texttt{max\_depth} & Integer $\in [3, 12]$ \\
 & \texttt{subsample} & Float $\in [0.5, 1.0]$ \\
 & \texttt{colsample\_bytree} & Float $\in [0.5, 1.0]$ \\
 & \texttt{gamma} & Float $\in [0.0, 5.0]$ \\
 & \texttt{min\_child\_weight} & Integer $\in [1, 10]$ \\
 & \texttt{reg\_alpha} ($L_1$) & Float $\in [0.0, 10.0]$ \\
 & \texttt{reg\_lambda} ($L_2$) & Float $\in [0.0, 10.0]$ \\ \midrule
\multirow{7}{*}{\textbf{MLP}} 
 & \texttt{n\_layers} & Integer $\in [0, 3]$ \\
 & \texttt{n\_units\_l\{i\}} & Integer $\in [32, 512]$, step $= 32$ \\
 & \texttt{dropout} & Float $\in [0.0, 0.5]$, step $= 0.1$ \\
 & \texttt{learning\_rate} (\texttt{lr}) & Log-Uniform Float $\in [10^{-5}, 10^{-2}]$ \\
 & \texttt{batch\_size} & Categorical $\in \{32, 64, 128\}$ \\
 & \texttt{epochs} & Integer $\in [10, 40]$ \\
 & \texttt{optimizer} & Categorical $\in \{\text{AdamW}, \text{RMSprop}, \text{SGD}\}$ \\ \midrule
\multirow{5}{*}{\textbf{Fine Tuning}} 
 & \texttt{hidden\_dim} (MLP Head) & Integer $\in [256, 1536]$, step  $= 128$ \\
 & \texttt{dropout} & Float $\in [0.0, 0.2]$, step $= 0.05$ \\
 & \texttt{learning\_rate} (\texttt{lr}) & Log-Uniform Float $\in [10^{-6}, 5\times 10^{-5}]$ \\
 & \texttt{learning\_rate\_head} (\texttt{lr}) & Log-Uniform Float $\in [10^{-4}, 5\times 10^{-3}]$ \\
 & \texttt{batch\_size} & Categorical $\in \{8, 16\}$ \\
 & \texttt{epochs} & Integer $\in [5, 15]$ \\ 
 & \texttt{frozen\_epochs} & Integer $\in [1, 3]$ \\ 
 & \texttt{warmup\_percentage} & Float $\in [0.0, 0.2]$, step $= 0.05$ \\ 
 & \texttt{pos\_weight\_alpha} & Float $\in [0.0, 1.0]$, step $= 0.1$ \\ 
 & \texttt{max\_grad\_norm} & Float $\in [0.2, 1.0]$, step $= 0.1$ \\ 
 \bottomrule
\end{tabular}%
\end{table}

A total of 30,350 hyperparameter configurations were evaluated. The five top-performing configurations, ranked by their $F1_{micro}$ score on the validation set, are listed in \cref{tab:top5_hyperparams}. The remaining hyperparameter configurations evaluated can be found in the Optuna results hosted within the project repository\footnote{\url{https://codeberg.org/JorgeDuenasLerin/psy-mapping-cie}}.

\begin{table}[ht]
\centering
\footnotesize
\caption{Optimal hyperparameters for the five highest-performing configurations based on validation $F1_{micro}$.}
\label{tab:top5_hyperparams}
\begin{tabular}{p{3.5cm}p{4cm}p{4cm}}
\toprule
Configuration & Parameter & Value \\
\midrule
\multirow{10}{*}{finetune\_e5\_large} & hidden\_dim & 1280 \\
 & dropout & 0.1 \\
 & lr & $1.5243337984464924 \times 10^{-5}$ \\
 & lr\_head & 0.000175389640525079 \\
 & batch\_size & 8 \\
 & epochs & 15 \\
 & frozen\_epochs & 1 \\
 & warmup\_percentage & 0.05 \\
 & pw\_alpha & 0.0 \\
 & max\_grad\_norm & 1.0 \\
\midrule
\multirow{9}{*}{mlt\_xgboost\_e5\_large} & n\_estimators & 759 \\
 & learning\_rate & 0.06812461682319591 \\
 & max\_depth & 9 \\
 & subsample & 0.5810888790819708 \\
 & colsample\_bytree & 0.6884800605089568 \\
 & gamma & 0.6242871360525342 \\
 & min\_child\_weight & 10 \\
 & reg\_alpha & 1.2494601393943636 \\
 & reg\_lambda & 6.7767174069276574 \\
\midrule
\multirow{9}{*}{mlt\_xgboost\_bio\_lord} & n\_estimators & 474 \\
 & learning\_rate & 0.06814228540442097 \\
 & max\_depth & 9 \\
 & subsample & 0.6899201652382992 \\
 & colsample\_bytree & 0.6088435881640017 \\
 & gamma & 1.4325774737625496 \\
 & min\_child\_weight & 4 \\
 & reg\_alpha & 0.5647227880560319 \\
 & reg\_lambda & 4.147124014825858 \\
\midrule
\multirow{9}{*}{mlt\_xgboost\_e5\_base} & n\_estimators & 684 \\
 & learning\_rate & 0.08897215138919123 \\
 & max\_depth & 8 \\
 & subsample & 0.7884814838592877 \\
 & colsample\_bytree & 0.8828431142426094 \\
 & gamma & 0.24276864122614503 \\
 & min\_child\_weight & 10 \\
 & reg\_alpha & 2.756079736396261 \\
 & reg\_lambda & 0.6687189811925814 \\
\midrule
\multirow{9}{*}{\shortstack[l]{mlt\_xgboost\\paraphrase\_multilingual}} & n\_estimators & 484 \\
 & learning\_rate & 0.11167218746749885 \\
 & max\_depth & 7 \\
 & subsample & 0.7860690412680696 \\
 & colsample\_bytree & 0.9280614811263148 \\
 & gamma & 0.23647686141974913 \\
 & min\_child\_weight & 6 \\
 & reg\_alpha & 1.1853215759117572 \\
 & reg\_lambda & 7.951972267900953 \\
\bottomrule
\end{tabular}
\end{table}

\subsection{Results}

The results presented in \cref{tab:results} reveal a broad spectrum of performance across the evaluated architectures, with $F1_{micro}$ scores ranging from a low of 0.245 for the \texttt{Doc2Vec} and \texttt{Random Forest} combination to a peak of 0.861 achieved by the \texttt{e5\_large} model paired with XGBoost. When analyzing text representation paradigms, it is evident that state-of-the-art \glspl{llm} embeddings, specifically \texttt{e5\_large}, \texttt{bio\_lord}, and \textit{paraphrase\_multilingual}, consistently outperform traditional and dimensionality reduction techniques. These transformer-based models effectively capture the complex semantic relationships and nuanced medical terminology required to describe psychiatric diagnoses.

Surprisingly, frequency-based methods like \gls{bow} and \acrfull{tfidf} remain highly competitive when processed by a \gls{mlp}, achieving $F1_{micro}$ scores near 0.84. This suggests that specific medical keywords within the diagnostic descriptions are highly discriminative even without deep contextual modeling. Conversely, \texttt{Doc2Vec} and \gls{lda} show the weakest performance, indicating that these methods struggle to generate high-quality dense representations for this specialized clinical domain compared to pre-trained transformer models.

Regarding the classification models, \texttt{XGBoost} emerges as the most robust choice for handling dense contextual embeddings, frequently yielding the highest $F1_{micro}$ and $F1_{macro}$ scores across the majority of \gls{llm} categories. In contrast, the \gls{mlp} demonstrates a unique affinity for high-dimensional sparse data, outperforming tree-based ensembles specifically when using \gls{bow} or \gls{tfidf} features. \texttt{Random Forest} tends to exhibit a more conservative classification behavior; while it achieves the highest precision scores in the study, peaking at 0.959 with \textit{e5\_large}, it suffers from significantly lower recall, which implies that it frequently fails to assign codes to valid diagnostic descriptions.

Finally, the notable discrepancy between $F1_{micro}$ and $F1_{macro}$ across all configurations underscores the persistent challenge posed by the long-tail distribution and severe class imbalance inherent in the medical coding of psychiatric disorders.

\begin{table}[ht]
\centering
\footnotesize
\caption{Performance metrics for all proposed text representations and classification models.}
\label{tab:results}
\renewcommand{\arraystretch}{0.95}
\setlength{\tabcolsep}{4pt}

\begin{tabular}{@{}llccccc@{}}
\toprule
\multirow{2}{*}{\textbf{Text Representation}} & \multirow{2}{*}{\textbf{Classification Model}} & \textbf{Validation} & \multicolumn{4}{c}{\textbf{Test}} \\
 & & \textbf{$F1_{micro}$} & \textbf{$F1_{micro}$} & \textbf{$F1_{macro}$} & \textbf{$Precision_{micro}$} & \textbf{$Recall_{micro}$} \\ \midrule

\multirow{3}{*}{BOW} & Random Forest & 0.571083 & 0.580911 & 0.321984 & 0.915951 & 0.425332 \\
  & XGBoost & 0.743010 & 0.763289 & 0.577667 & 0.913800 & 0.655347 \\
  & MLP & 0.838915 & 0.841418 & 0.760053 & 0.915043 & 0.778759 \\
 \midrule

\multirow{3}{*}{TFIDF} & Random Forest & 0.573161 & 0.581705 & 0.316725 & 0.913839 & 0.426642 \\
  & XGBoost & 0.798111 & 0.809734 & 0.713034 & 0.933889 & 0.714717 \\
  & MLP & 0.839446 & 0.843500 & 0.771153 & 0.921017 & 0.778018 \\
 \midrule

\multirow{3}{*}{LSA} & Random Forest & 0.655926 & 0.673023 & 0.433522 & 0.950900 & 0.520825 \\
  & XGBoost & 0.843108 & 0.847121 & 0.745676 & 0.939482 & 0.771295 \\
  & MLP & 0.810317 & 0.820342 & 0.719420 & 0.901967 & 0.752265 \\
 \midrule

\multirow{3}{*}{LDA} & Random Forest & 0.520411 & 0.514215 & 0.277046 & 0.809973 & 0.376674 \\
  & XGBoost & 0.629881 & 0.655024 & 0.437316 & 0.846300 & 0.534272 \\
  & MLP & 0.582709 & 0.567703 & 0.344954 & 0.754106 & 0.455188 \\
 \midrule

\multirow{3}{*}{Doc2Vec} & Random Forest & 0.257799 & 0.245057 & 0.078746 & 0.437527 & 0.170190 \\
  & XGBoost & 0.494931 & 0.503974 & 0.265580 & 0.842773 & 0.359467 \\
  & MLP & 0.544382 & 0.565097 & 0.348673 & 0.716750 & 0.466412 \\
 \midrule

\multirow{3}{*}{bert\_spanish} & Random Forest & 0.627271 & 0.635806 & 0.485102 & 0.841212 & 0.511025 \\
  & XGBoost & 0.845527 & 0.843151 & 0.757935 & 0.933693 & 0.768617 \\
  & MLP & 0.668457 & 0.668743 & 0.552860 & 0.823785 & 0.562817 \\
 \midrule

\multirow{3}{*}{bio\_lord} & Random Forest & 0.816930 & 0.814334 & 0.705052 & 0.940584 & 0.717965 \\
  & XGBoost & 0.860543 & 0.856568 & 0.774323 & 0.937415 & 0.788559 \\
  & MLP & 0.798751 & 0.804804 & 0.718908 & 0.894909 & 0.731183 \\
 \midrule

\multirow{3}{*}{bsc\_bio} & Random Forest & 0.732788 & 0.731295 & 0.559054 & 0.945680 & 0.596148 \\
  & XGBoost & 0.848944 & 0.850545 & 0.766418 & 0.940312 & 0.776423 \\
  & MLP & 0.761383 & 0.783672 & 0.689200 & 0.903564 & 0.691869 \\
 \midrule

\multirow{3}{*}{e5\_base} & Random Forest & 0.811167 & 0.807350 & 0.703085 & 0.958787 & 0.697225 \\
  & XGBoost & 0.860422 & 0.859394 & 0.765789 & 0.938715 & 0.792433 \\
  & MLP & 0.763853 & 0.767765 & 0.657483 & 0.902672 & 0.667939 \\
 \midrule

\multirow{3}{*}{e5\_large} & Random Forest & 0.818014 & 0.816224 & 0.717595 & \textbf{0.959221} & 0.710330 \\
  & XGBoost & 0.861704 & \textbf{0.861687} & 0.757278 & 0.938684 & \textbf{0.796365} \\
  & MLP & 0.763060 & 0.775304 & 0.663071 & 0.920439 & 0.669705 \\
 \midrule

\multirow{3}{*}{llama3\_8b} & Random Forest & 0.757071 & 0.750696 & 0.642823 & 0.929732 & 0.629480 \\
  & XGBoost & 0.843395 & 0.837823 & 0.748553 & 0.927687 & 0.763831 \\
  & MLP & 0.779353 & 0.787452 & 0.703356 & 0.903259 & 0.697966 \\
 \midrule

\multirow{3}{*}{paraphrase\_multilangual} & Random Forest & 0.811714 & 0.806881 & 0.708714 & 0.949418 & 0.701555 \\
  & XGBoost & 0.859050 & 0.857787 & \textbf{0.779577} & 0.936720 & 0.791123 \\
  & MLP & 0.758207 & 0.772784 & 0.673610 & 0.903792 & 0.674947 \\
 \midrule

\multirow{3}{*}{plantl\_roberta} & Random Forest & 0.788714 & 0.783272 & 0.677993 & 0.937073 & 0.672839 \\
  & XGBoost & 0.848140 & 0.846480 & 0.773269 & 0.927263 & 0.778645 \\
  & MLP & 0.774670 & 0.785358 & 0.692220 & 0.895701 & 0.699219 \\
 \bottomrule

\end{tabular}
\end{table}

Among all evaluated fixed-representation configurations, the \texttt{e5\_large} embedding model consistently achieved the highest test $F1_{micro}$ scores, outperforming both classical representations and other transformer-based embeddings. Based on this result, \texttt{e5\_large} was selected as the sole backbone for transformer fine-tuning, as the significant computational cost and high resource requirements associated with updating all internal weights of such large-scale models made it impractical to perform end-to-end training for every candidate. This transition from static feature extraction to active classification allowed the architecture to adapt its deep linguistic representations specifically to the medical nomenclature and the unique long-tail distribution of psychiatric codes. 

Ultimately, this approach positioned \texttt{e5\_large} as the overall best-performing model in our study. Under this optimized strategy, the model achieved a validation $F1_{micro}$ of 0.862164, reaching the highest performance levels observed. Specifically, the fine-tuned model yielded a test $F1_{micro}$ of 0.866408 and an $F1_{macro}$ of 0.804303, supported by a precision micro of 0.910220 and a recall micro of 0.826620.

\begin{figure}[ht]
    \centering
    \includegraphics[width=0.95\textwidth]{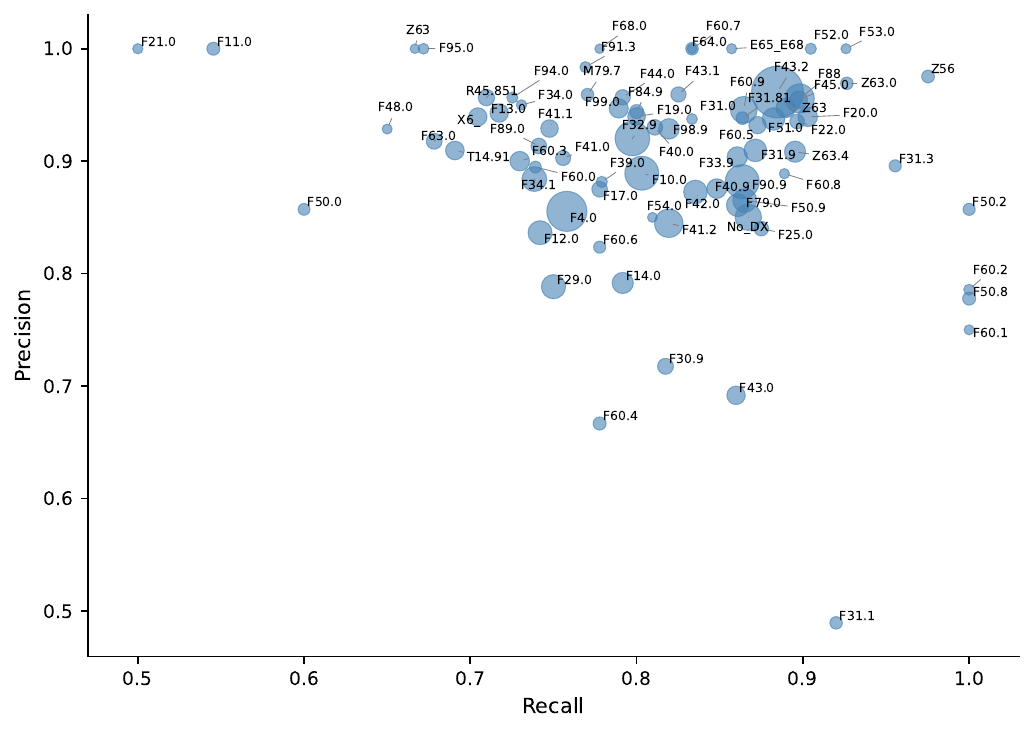}
    \caption{Per-class precision vs.\ recall. Bubble size proportional to number of samples. Four classes with zero precision are omitted (see text).}
    \label{fig:precision_recall}
\end{figure}

To provide a granular evaluation of the model's performance beyond global metrics, a per-class analysis was conducted, as the high number of \gls{icd} codes can mask individual variations in predictive quality. This granular perspective is visualized in \cref{fig:precision_recall}, where the precision and recall for each category are plotted against each other, with bubble sizes proportional to the number of samples. The visualization reveals a clear trend where classes with a higher number of samples—represented by the larger bubbles—tend to achieve a superior and more stable balance between precision and recall, clustering firmly in the upper-right quadrant. In contrast, the increased variance observed in smaller bubbles highlights the challenges of classifying rarer conditions, including the four minority classes (F32.0 with 1 sample, F32.3 with 55 samples, F33.0 with 2 samples, and F65.0 with 10 samples) that were omitted from the figure due to a total lack of predictive success.

\section{Discussion} \label{sec:discussion}
The comparative analysis presented herein illuminates the critical trade-offs between computational efficiency and semantic depth in automated psychiatric coding. While our results quantitatively demonstrate the superiority of transformer-based embeddings over classical \gls{bow} approaches, the qualitative interpretation of these findings offers deeper insights into the nature of clinical language processing within the Spanish medical context. 

The consistent outperformance of \gls{llm}-derived embeddings (\textit{e5\_large}, \textit{bio\_lord}) over \gls{bow} and \gls{tfidf} methods supports the hypothesis that psychiatric diagnoses often rely on implicit semantic cues rather than explicit keyword matching. Psychiatric notes frequently employ subjective descriptions (e.g., ``feeling empty'', ``loss of motivation'') which do not always map directly to \gls{icd} terminology. Frequency-based models, while achieving competitive $F1_{micro}$ scores due to their sensitivity to high-frequency terms like \textit{depression} or \textit{anxiety}, fail to generalize when these symptoms are described euphemistically or contextually. This aligns with prior findings in clinical \gls{nlp} regarding the ``lexical gap'' between patient narratives and standardized taxonomies \cite{silva2024aiding}.

The failure of \texttt{Doc2Vec} and \gls{lda} further suggests that unsupervised topic modeling on general corpora cannot capture the idiosyncratic syntax of Spanish psychiatric documentation, reinforcing the necessity for domain-specific pretraining in this linguistic space.

The divergence in performance between \textit{XGBoost} and \glspl{mlp} based on feature type offers valuable architectural guidance. The robustness of \texttt{XGBoost} when processing dense embeddings may be attributed to its inherent regularization capabilities, which prevent overfitting on the high-dimensional latent spaces produced by transformers. Conversely, the \glspl{mlp} superior performance with sparse \gls{bow}/\gls{tfidf} features likely stems from its ability to learn complex non-linear interactions between keyword co-occurrences without the dimensionality reduction penalties that tree ensembles might impose on sparse data. This suggests that model selection should not be decoupled from feature extraction; a hybrid approach where the classifier architecture is optimized jointly with the representation paradigm could yield further marginal gains, though at increased computational cost.

Despite the semantic advantages of \glspl{llm}, the discrepancy between $F1_{micro}$ and $F1_{macro}$ scores remains a significant concern. The fact that certain rare classes (e.g., F32.0, F65.0) yielded zero precision even with fine-tuned models indicates that semantic richness alone cannot overcome data scarcity. In clinical settings, this translates to a risk of systematic under-coding for less common but potentially critical conditions. The reliance on supervised learning exacerbates this issue; without sufficient positive examples during training, the model defaults to predicting majority classes to minimize loss. This highlights a fundamental limitation in current automated coding systems: they are optimized for administrative efficiency (coding frequent disorders) rather than clinical comprehensiveness.

\section{Conclusions and Future Work} \label{sec:conclusions}
This study presents a comprehensive evaluation of automated \gls{icd} classification for psychiatric diagnoses in Spanish, bridging the gap between classical \gls{nlp} techniques and modern \gls{llm} architectures. By systematically benchmarking text representation paradigms and classification models on a dataset of 79,048 clinical entries, we have established that domain-specific contextual embeddings significantly outperform traditional frequency-based methods in capturing the semantic complexity of psychiatric discourse.

Our primary findings confirm that transformer-based embeddings, specifically \textit{e5\_large}, provide the most robust foundation for automated coding tasks in this domain. When paired with optimized classifiers, these representations achieve a test $F1_{micro}$ score of $0.861687$ and a $F1_{macro}$ score of $0.757278$. This performance surpasses classical \gls{bow} and \gls{tfidf} approaches, validating the hypothesis that semantic nuance is critical for distinguishing between clinically similar psychiatric conditions. 

Furthermore, our analysis reveals distinct synergies between feature types and classifier architectures: \texttt{XGBoost} excels with dense embeddings due to its regularization properties, while \glspl{mlp} remain competitive for sparse keyword-based features. The successful fine-tuning of \textit{e5\_large} (achieving a test $F1_{micro}$ score of $0.866408$ and a $F1_{macro}$ score of $0.804303$) demonstrates that task-specific adaptation can effectively mitigate the challenges posed by the long-tail distribution of diagnostic codes without incurring prohibitive computational costs.

Future research directions should prioritize enhancing system explainability through interpretability tools such as attention visualization, which will provide evidence trails for code assignments and foster clinical trust alongside auditability requirements. Concurrently, efforts must address data scarcity by developing specialized techniques like synthetic data generation or focal loss to improve recall for rare but high-risk diagnoses that currently suffer from underrepresentation. To facilitate broader adoption, future work should also investigate deployment optimization strategies including knowledge distillation and quantization, enabling the real-time inference of large models within resource-constrained hospital environments without significant performance loss. Finally, expanding the system's scope through multimodal integration would allow for the incorporation of structured clinical data alongside textual records to resolve diagnostic ambiguities more effectively than unstructured text alone.

\section*{Acknowledgements}
This study is supported by the Dirección General de Investigación e Innovación Tecnológica de la Comunidad de Madrid (Orden 3177/2024) through the I+D Technological activities program (TEC-2024/COM-224); CIBER -Consorcio Centro de Investigación Biomédica en Red- (CB/07/09/0025); the Instituto de Salud Carlos III with the support of the European Regional Development Fund (ISCIII PI23/00614; PMP24/00026); Fundació la Marató de TV3 (202226-31) and by CaixaResearch Health 2023 LCF/PR/HR23/52430033.

\bibliographystyle{unsrtnat}
\bibliography{cas-refs}

\appendix
\section{List of Mental Health \gls{icd} Codes} \label{apx:cie-10-codes}

\Cref{tab:cie10_full_list} provides the complete list of the 85 diagnostic categories used in this study, including both standard \gls{icd} codes and internal project identifiers.

\begin{small}
\begin{longtable}{lp{0.75\textwidth}}
\caption{Selected diagnostic codes and their corresponding English descriptions.} \label{tab:cie10_full_list} \\
\toprule
\textbf{Code} & \textbf{Description} \\ \midrule
\endfirsthead

\multicolumn{2}{c}{{\bfseries \tablename\ \thetable{} -- Continued from previous page}} \\
\toprule
\textbf{Code} & \textbf{Description} \\ \midrule
\endhead

\midrule
\multicolumn{2}{r}{{Continued on next page}} \\
\bottomrule
\endfoot

\bottomrule
\endlastfoot

F30.9 & Manic episode, unspecified \\
F31.0 & Bipolar affective disorder, current episode hypomanic \\
F31.1 & Bipolar affective disorder, current episode manic without psychotic symptoms \\
F31.3 & Bipolar affective disorder, current episode mild or moderate depression \\
F31.81 & Bipolar II disorder \\
F31.9 & Bipolar affective disorder, unspecified \\
F32.9 & Depressive episode, unspecified \\
F33.9 & Recurrent depressive disorder, unspecified \\
F34.0 & Cyclothymia \\
F34.1 & Dysthymia \\
F39 & Unspecified mood [affective] disorder \\
F40.9 & Phobic anxiety disorder, unspecified \\
F40.0 & Agoraphobia \\
F41.0 & Panic disorder [episodic paroxysmal anxiety] \\
F41.1 & Generalized anxiety disorder \\
F41.2 & Mixed anxiety and depressive disorder \\
F42 & Obsessive-compulsive disorder \\
F43 & Reaction to severe stress, and adjustment disorders \\
F43.1 & Post-traumatic stress disorder \\
F43.2 & Adjustment disorder \\
F44 & Dissociative [conversion] disorder \\
F45 & Somatoform disorder \\
F48 & Other neurotic disorders (chronic fatigue, depersonalization-derealization) \\
F50.9 & Eating disorder, unspecified \\
F50.0 & Anorexia nervosa \\
F50.2 & Bulimia nervosa \\
F50.8 & Other eating disorders \\
F51 & Nonorganic sleep disorders \\
F51.0 & Nonorganic insomnia \\
F52 & Sexual dysfunction not caused by organic disorder or disease \\
F53 & Mental and behavioral disorders associated with the puerperium \\
F54 & Psychological and behavioral factors associated with disorders or diseases classified elsewhere \\
F60.0 & Paranoid personality disorder \\
F60.1 & Schizoid personality disorder \\
F60.2 & Dissocial personality disorder \\
F60.3 & Emotionally unstable personality disorder \\
F60.4 & Histrionic personality disorder \\
F60.5 & Anankastic personality disorder \\
F60.6 & Anxious [avoidant] personality disorder \\
F60.7 & Dependent personality disorder \\
F63.0 & Pathological gambling \\
F60.8 & Other specific personality disorders \\
F60.9 & Personality disorder, unspecified \\
F64 & Gender identity disorders \\
F65 & Disorders of sexual preference: Paedophilia, paraphilia \\
F68 & Other disorders of adult personality and behavior (factitious) \\
F79 & Unspecified mental retardation \\
F84.9 & Autism spectrum disorder \\
F89 & Unspecified disorder of psychological development \\
F90.9 & Hyperkinetic disorder, unspecified \\
F91.3 & Oppositional defiant disorder \\
F94 & Disorders of social functioning with onset specific to childhood \\
F95 & Tic disorder \\
F98.9 & Unspecified behavioral and emotional disorders with onset usually occurring in childhood \\
F99 & Mental disorder, not otherwise specified \\
R45.851 & Suicidal ideation \\
T14.91 & Suicide attempt \\
X6\_ & Intentional self-poisoning \\
M79.7 & Fibromyalgia \\
Z63 & Other problems related to primary support group, including family circumstances \\
Z63.4 & Problems related to disappearance or death of family member \\
E65\_E68 & Obesity and other hyperalimentation \\
F10 & Mental and behavioral disorders due to use of alcohol \\
F11 & Mental and behavioral disorders due to use of opioids \\
F12 & Mental and behavioral disorders due to use of cannabinoids \\
F13 & Mental and behavioral disorders due to use of sedatives or hypnotics \\
F14 & Mental and behavioral disorders due to use of cocaine \\
F17 & Mental and behavioral disorders due to use of tobacco \\
F19 & Mental and behavioral disorders due to multiple drug use and use of other psychoactive substances \\
F20 & Schizophrenia \\
F21 & Schizotypal disorder \\
F22 & Persistent delusional disorders \\
F25 & Schizoaffective disorders \\
F29 & Unspecified nonorganic psychosis \\
F32.3 & Depressive episode with psychotic symptoms \\
F32 & Depressive episode \\
F33 & Recurrent depressive disorder \\
F4 & Anxiety, dissociative, stress-related, somatoform and other nonpsychotic mental disorders \\
F40 & Phobic anxiety disorders \\
F50 & Eating disorders \\
F88 & Cognitive dimension \\
Z63 & Family problems and conflicts \\
Z56 & Work-related problems and conflicts \\
No\_DX & No diagnosis \\
Z63.0 & Relationship problems and conflicts \\
\end{longtable}
\end{small}

\end{document}